%% file: Self-Reorganizing_and_Rejuvenating_CNNs_for_Increasing_Model_Capacity_Utilization.tex
\documentclass[10pt,twocolumn,letterpaper]{article}

\usepackage{cvpr}
\usepackage{times}
\usepackage{epsfig}
\usepackage{graphicx}
\usepackage{amsmath}
\usepackage{amssymb}

\usepackage{booktabs}       

\usepackage{multirow}
\usepackage{float}
\usepackage[ruled]{algorithm2e}
\usepackage{amsmath}
\usepackage{caption}	
\usepackage{enumitem}
\usepackage[english]{babel}
\usepackage{graphicx}
\usepackage{algorithmic}

\usepackage[breaklinks=true,bookmarks=false]{hyperref}

\cvprfinalcopy 

\begin{document}

\title{Self-Reorganizing and Rejuvenating CNNs for Increasing Model Capacity Utilization}

\author{Wissam J. Baddar, Seungju Han, Seonmin Rhee, Jae-Joon Han\\
Samsung Advanced Institute of Technology (SAIT), South Korea\\
{\tt\small wisam.baddar@samsung.com, sj75.han@samsung.com, s.rhee@samsung.com, jae-joon.han@samsung.com}
}

\maketitle


\begin{abstract}
	\input{./sections/00-abs.tex}
\end{abstract}

\label{sec:intro}
\section{Introduction}
	\input{./sections/01-intro.tex}

\section{Proposed Self-Reorganizing and Rejuvenating CNN}
	\input{./figs/fig1.tex}

\input{./sections/03-00-proposed-overview.tex}

	\subsection{Notation}
	\input{./sections/03-00-notation.tex}

	\subsection{Class Representative Activation}
		\input{./sections/03-01-class-representative-features.tex}

	\subsection{Self-Reorganization}
		\input{./sections/03-02-neuron-reorganization.tex}
	
	\subsection{Neuron Rejuvenation}

\input{./sections/03-03-neuron-rejuvenation.tex}

	\subsection{Training Procedure}
		\input{./alg/alg_01.tex}		
		\input{./sections/03-04-training_strategy.tex}

\section{Experiments}
	\input{./sections/04-00-exp.tex}
	
	\subsection{Ablation Study}
		\subsubsection{Effect of Self-Reorganization \& Rejuvenation on different layers}
		\input{./sections/04-01-exp1_1.tex}
		\input{./tables/table_01.tex}

		\subsubsection{Effectiveness of the proposed self-reorganization}
		\input{./sections/04-01-exp1_2.tex}

		\input{./tables/table_02.tex}

	\subsection{Comparative Experiments}
		\input{./tables/table_03.tex}

		\input{./tables/table_04.tex}

		\input{./figs/fig2.tex}
		\input{./sections/04-02-exp2.tex}

	\subsection{Visualizations of Self-Reorganized \& Rejuvenated CNNs}
		\input{./sections/04-03-exp3.tex}

\section{Conclusion}
	\input{./sections/05-conclusion.tex}

{\small
\bibliographystyle{ieee_fullname}
\bibliography{ref}
}

\section{Appendix}
	\subsection{Self-organizing feature maps (SOFM)}
		\input{./sections/Appendix-SOFM.tex}

		\input{./alg/alg_02.tex}

\end{document}

%% file: sections/00-abs.tex
In this paper, we propose self-reorganizing and rejuvenating convolutional neural networks; a biologically inspired method for improving the computational resource utilization of neural networks. The proposed method utilizes the channel activations of a convolution layer in order to reorganize that layers parameters. The reorganized parameters are clustered to avoid parameter redundancies. As such, redundant neurons with similar activations are merged leaving room for the remaining parameters to rejuvenate. The rejuvenated parameters learn different features to supplement those learned by the reorganized surviving parameters. As a result, the network capacity utilization increases improving the baseline network performance without any changes to the network structure. The proposed method can be applied to various network architectures during the training stage, or applied to a pre-trained model improving its performance. Experimental results showed that the proposed method is model-agnostic and can be applied to any backbone architecture increasing its performance due to the elevated utilization of the network capacity.

%% file: sections/01-intro.tex
Deep neural networks, particularly convolutional neural networks (CNNs), have shown state-of-the-art performance in multiple applications. Impressive performances in applications like image classification \cite{resnet,densenet,efficientnet,nasnet}, object detection \cite{FastRCNN,FasterRCNN,deepid,yolov3}, or even segmentation \cite{deepmask,Sharpmask,unet,dcan,FCIS} have been achieved due to the rapid development of powerful hardware, which made it easier to train complicated CNN models with large capacities. In fact, one can observe that for large-scale tasks such as ImageNet classification \cite{imagenet}, models with a larger number of parameters (deeper and wider models) have been achieving state-of-the-art performances \cite{NR}.

Utilizing models with large number of parameters is consistent with the intuition that highly non-linear relationships between input features and outputs require very expressive models with sufficient capacity. In fact, in many cases, networks were purposefully over-parametrized to win the filter lottery tickets \cite{lottery}. However, increasing model sizes makes them vulnerable to capturing noise rather than the intended patterns \cite{dsd}. Moreover, employing larger models increases the number of redundant parameters and reduces the efficiency of which these models utilize their capacities. Indeed, previous work have shown that many SGD trained neural network models are severely underutilized \cite{netslimming,lottery}. This problem is magnified when models that are known to work well on large-scale tasks are utilized to solve smaller tasks. For example, the authors of \cite{netslimming} have shown that a VGG network \cite{vgg} can be pruned by a factor of 10 without changing the accuracy on the CIFAR-100 dataset \cite{cifar100}. On the contrary, simply reducing the model capacity could lead to  models that under-fit the task at hand. For example, it has been widely accepted to sacrifice the model performance for the sake of reducing the models complexity \cite{netslimming,prune1,prune2,prune3}. Popular methods that have been proposed to tackle the problem of over-parameterization in CNNs can be divided into the following categories: 

\textbf{Architecture search methods:} \cite{AS1,AS2,AS3,AS4,nasnet} These methods try to search for an architecture in a pre-defined architectures space. To be able to do so, they need to train a multitude of networks to finally obtain the optimal network architecture. Neural architecture search requires multiple training passes, which is extremely computational expensive. It should be noted that although neural architecture search could find an architecture that optimizes the performance, it does not guaranty that the model is not over-parametrized, or contains a large number of redundancies. 

\textbf{Model pruning and compression:} \cite{netslimming,prune1,prune2,prune3} These methods try to reduce the computational complexity of a model by removing unnecessary connections or nodes from the network. By doing so, the number of parameters of a pre-trained model are reduced. In other words, the search-space for these methods exists within the capacity of the pre-trained model. The majority of pruning-based methods require a certain criterion which identifies the least relevant connections in order to prune them out. However, finding the pruning criterion can be a challenging task. Added to that, the reduction of model complexity often comes on the expense of performance. It should be also noted that, these methods endeavor to eliminate the irrelevant parameters but does not try to re-purpose them to expand the models capacity. 
  

\textbf{Expanding the baseline model capacity:} \cite{dsd,NR} Unlike pruning, these methods try to employ the irrelevant parameters to increase the model capacity and improve its performance. Dense-Sparse-Dense (DSD) \cite{dsd} proposed to sparsify an over-parametrized network  via pruning with the smallest-norm-less-informative assumption. The pruned version of the network is fine-tuned. Finally, the pruned parameters are re-initialized and the model is fine-tuned to improve the model performance. However, this method cannot be applied to any model, especially when the smallest-norm-less-informative assumption is difficult to apply \cite{rethinking,geometric}. The work in \cite{NR} introduced a sparsity constraint on the standard deviation parameter ($\gamma$) of the batch normalization layer. Connections with a $\gamma$ parameters below a certain threshold were re-initialized and rejuvenated. However, defining the threshold for $\gamma$ can be very empirical. Added to that, if the layer does not have a batch normalization operation, the method in \cite{NR} cannot be applied.

In this paper, we propose a novel strategy to expand the capacity of any baseline model by rejuvenating redundant parameters and efficiently utilizing them. The proposed self-reorganizing and rejuvenating CNNs draws inspiration from the structure of the human visual cortex. In the visual cortex, lower abstraction levels are encoded in the primary visual cortex (V1). In V1, an abundance of neurons are fired sparsely across the V1 corresponding to simple stimuli that are simple constructs (e.g., edges, orientations and colors) \cite{V1,V1IT}. When the signal propagates up the ventral stream, activations of neurons start to get clustered into specific locations in the inferior temporal (IT) cortex that correspond to certain higher abstraction levels \cite{V1IT,IT}. To mimic the clustering of activations in the IT cortex, the proposed method employs self-organizing feature maps (SOFM) to reorganize the layer parameters such that clustered parameters respond to similar high abstraction stimuli. In other words, the proposed method relies on the activations of a layer in order to project the layer parameters into a compressed space. By doing so, the proposed method makes room for new features to be learned in the remaining parameters. The main advantages of the proposed method can be summarized in the following:

\begin{itemize}[leftmargin=*]


\item The proposed self-reorganizing and rejuvenating CNN improves the performance of a model by increasing the model capacity utilization without modifying its structure. To do so, the proposed method can be applied during the training of a model from scratch or it can be applied to a pre-trained model. 

\item The proposed method is model agnostic and can be applied to any CNN. Unlike methods that rely on pruning the model \cite{dsd,NR}, which require laborious parameter tuning, the proposed method aggregates parameters with similar activations into a compressed space. The method in \cite{dsd} could not be applied when the smallest-norm-less-informative assumption is unsatisfied \cite{rethinking,geometric}, and \cite{NR} requires a batch-normalization operation in order to be applied.

\item The proposed method mimics the human visual cortex by aggregating parameters with similar activations into a compressed space. Results showed that applying the proposed method to deeper layers of the model, the proposed method can improve the performance of shallow models to be comparable to deeper CNN models.
\end{itemize}

%% file: figs/fig1.tex
\begin{figure*}[!ht]
\begin{center}
	\includegraphics[width=1\textwidth,keepaspectratio]{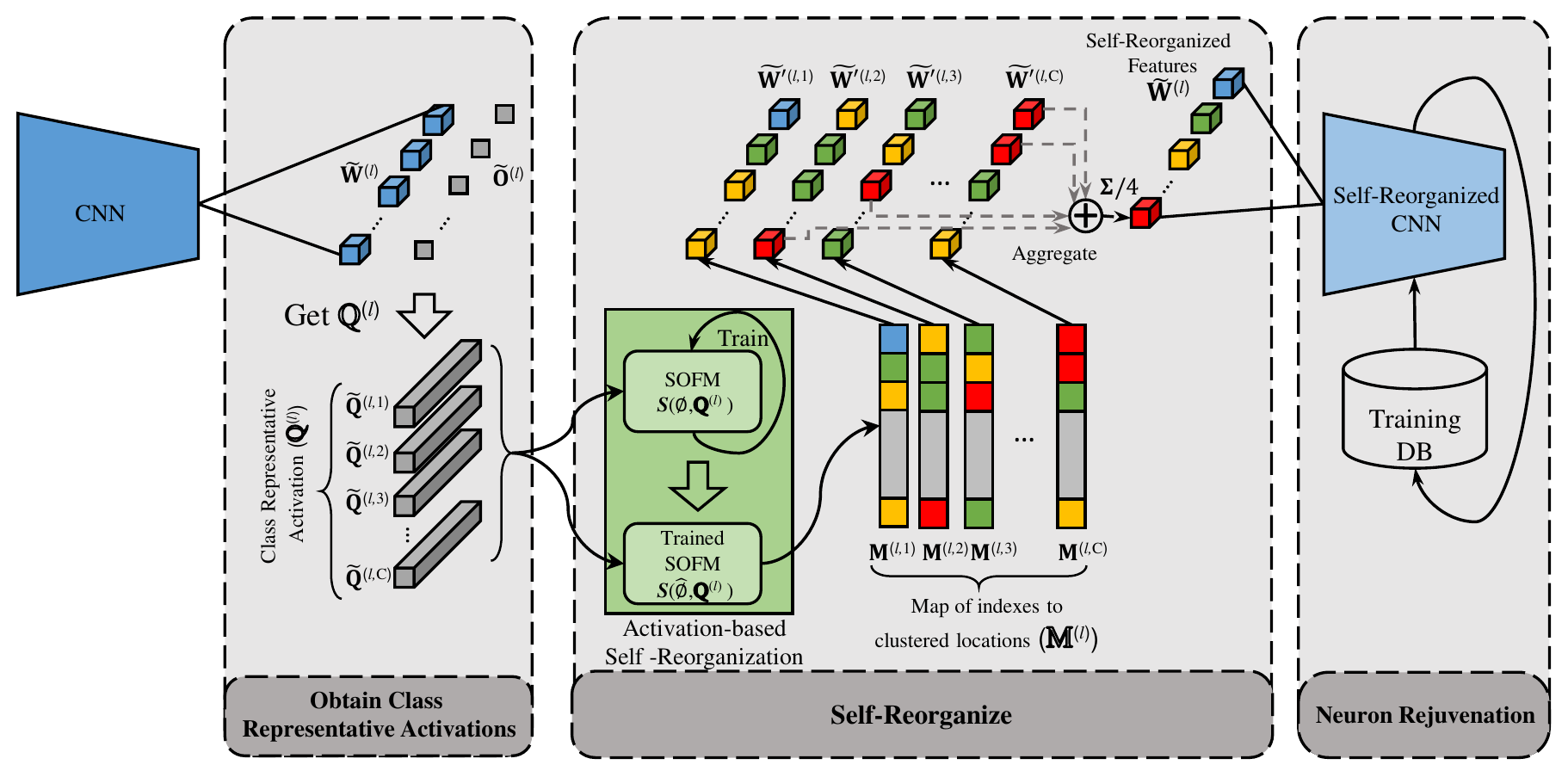}
\end{center}
   \caption{Overview of the proposed self-reorganizing and rejuvenating CNN. Best viewed in color.}
\label{fig:1}
\end{figure*}

%% file: sections/03-00-proposed-overview.tex
Figure~\ref{fig:1} shows an overview of the proposed self-reorganizing and rejuvenating CNN. The proposed method consists of three main steps; obtaining class representative activations, self-reorganization, and finally neuron rejuvenation. In the first step, a class representative activation for each class is obtained from a layer in the model. The class representative activations are obtained such that they represent the feature map of each output channel of that layer according to each class fed into the model. In the second step, the class representative activations are utilized to train a self-organizing feature map to cluster and compress the layer activations. The clustered activations are then utilized to generate a mapping to reorganize the layer parameters accordingly. After the weights are reorganized, both the reorganized parameters and the parametrs of non-surviving channels are rejuvenated. The details of each step of the proposed self-reorganizing and rejuvenating CNN are described in the following subsections.

%% file: sections/03-00-notation.tex
Before describing the details of the proposed method, we clarify the notion used hereafter. A bold faced capital symbol with a tilde (e.g., $\mathbf{\widetilde{W}}$) is used to represent tensors of rank 2 or higher. A vector is represented as a bold faced capital symbol (e.g., $\mathbf{W}$). A plain and small symbol, like $b$, represents a scalar. Finally, lower subscripts indicate indexing, while  superscript with parenthesis represent the network layer. For example, $\mathbf{\widetilde{W}}_{i,j}$ indicates the element at position $(i,j)$ of the 2D matrix $\mathbf{\widetilde{W}}$ and $\mathbf{\widetilde{W}}_{i,:,:,:}^{(l)}$ is the $i$th input channel of the weight parameter of layer $l$ of the network.

%% file: sections/03-01-class-representative-features.tex
Given a convolutional layer $l$ that we are interested in reorganizing and rejuvenating. The layer $l$ is parametrized by the weight tensor $\mathbf{\widetilde{W}}^{(l)} \in {\rm I\!R}^{i \times o \times w \times h}$ and the bias vector $\mathbf{B}^{(l)} \in {\rm I\!R}^{o}$, where $i,o,h,w$ are the number of input channels, output channels, height and width of the layers trainable tensor, respectively. The output activations of the layer $l$ can be described with the tensor $\mathbf{\widetilde{O}}^{(l)} \in {\rm I\!R}^{o \times w \times h}$. The set of class representative activation is represented by: 


\begin{equation}
    \mathbb{Q}^{(l)} = \left \{ \mathbf{\widetilde{Q}}^{(l,1)}, ... \mathbf{\widetilde{Q}}^{(l,C)}: \mathbf{\widetilde{Q}}^{(l,c)} = \dfrac{1}{N_c} \sum_{i=1}^{N_c} \mathbf{\widetilde{O}}_i^{(l)}  \right \}.
\end{equation}


$\mathbb{Q}^{(l)}$ is a set of 3-D tensors $\mathbf{\widetilde{Q}}^{(l,c)}$, each representing a class representative activation of a certain class $c$ at the layer $l$. In other words, $\mathbf{\widetilde{Q}}^{(l,c)}$ is the mean of all activations of the training samples of class $c$ at the layer $l$, where $N_c$ is the total number of training samples in class $c$. Notice that, $\mathbf{\widetilde{Q}}^{(l,c)} \in {\rm I\!R}^{o \times w \times h}$ and a concatenated version of $\mathbb{Q}^{(l)}$ is the tensor $\mathbf{\widetilde{Q}}^{(l)} \in {\rm I\!R}^{oC \times w \times h}$. Each slice of $\mathbf{\widetilde{Q}}^{(l,c)}_{o,:,:}$ is 2-D matrix representing a channel output activation of layer $l$ corresponding to one of the classes ($c=1,2,...,C$) in the training set.

%% file: sections/03-02-neuron-reorganization.tex
In this section, we detail how the class-representative activations ($\mathbb{Q}^{(l)}$) are utilized to reorganize the parameters of a certain layer $l$. The reorganization of the features can be achieved by clustering and rearranging the trainable parameters of layer $l$ (i.e., weights $\mathbf{\widetilde{W}}$, biases $\mathbf{B}$, and batch-norm mean ($\beta$) and variance ($\gamma$ ) parameters). Influenced by the IT cortex, we propose clustering neuron parameters according to the similarity between output channel activations of that layer with respect to different inputs from previous layers. In other words, clustering is performed by considering the similarities between the class-representative activation slices $(\mathbf{\widetilde{Q}}^{(l,c)}_{o,:,:})$. For the clustering, a self reorganizing feature map (SOFM) is utilized.

SOFM $(S(\phi,\mathbb{Q}^{(l)}))$ was introduced by Teuvo Kohonen in \cite{SOM1,SOM2}. SOFM is a type of artificial neural network that is trained using unsupervised learning to produce a low-dimensional map of the input space of the training samples. SOFM is a stochastic algorithm that updates the weights of the neurons at each step. However, instead of updating weights with error-correction learning, training the SOFM relies on competitive learning. The reasons for selecting SOFM as the clustering method in our work is two fold: First, the competitive learning algorithm in SOFM utilizes a neighborhood function to preserve the topological properties of the input space. This means that the map of the SOFM morphs into a clustered representation of the training samples, such that the geometric distance between the map nodes corresponds to the similarity between those nodes and the training samples $(\mathbf{\widetilde{Q}}^{(l,c)}_{o,:,:})$. This allows us to reorganize the location of the nodes (channel parameters) of the layer $l$ according the output activation of those channels. Second, The SOFM does not require the number of clusters to be known before-hand. This is useful for our case, given that the number of clusters in the activations obtained from the convolution layer $l$ are unknown. 

In this work, the SOFM $(S(\phi,\mathbb{Q}^{(l)}))$ has a 1-D trainable feature map $\phi$. Each parameter $(\phi_i)$ in the map is of dimension $w \times h$, which is the size of flattened slice of the class-representative activation ($\mathbf{\widetilde{Q}}^{(l,c)}_{o,:,:}$). To train the SOFM $(S(\phi,\mathbb{Q}^{(l)}))$ with the class-representative activation, each slice of the class-representative activation ($\mathbf{\widetilde{Q}}^{(l,c)}_{o,:,:}$) is used as a single sample. Note that, the length of the 1-D map of SOFM $(\phi_i)$ can be less than or equal to the number of output channels $(o)$ of the layer $l$. When the length of $(\phi_i)$ is equal to the number of output channels ($o$), the parameters of $l$ are reorganized into the same number of filters in $l$. However, when the length of $\phi_i$ is less than $o$, reorganization is coupled with a compression of the parameters into a smaller number of filters. In this case, the remaining filters are then reinitialize in the rejuvenation process. Details on the training process of the SOFM are included in the supplementary material.
 

As aforementioned, the feature map $(\widehat{\phi})$ of the trained SOFM $(S(\widehat{\phi},\mathbb{Q}^{(l)}))$, is a clustered version of the training samples. Hence, feeding a sample to the SOFM $(S(\widehat{\phi},\mathbf{\widetilde{Q}}^{(l,c)}_{o,:,:}))$ would provide an index to where that sample belongs on the feature map $(\widehat{\phi})$. Note that, we do not need to know the number of clusters on that features map, we are only interested in the indexing of the samples on that features map. To reorganize the parameters of layer $l$, we first obtain the set of mapping indexes using $(S(\widehat{\phi},\mathbb{Q}^{(l)}))$ . The map of indexes to the clustered locations is obtained as follows:

\begin{equation}
    \mathbb{M}^{(l)} = \left \{ \mathbf{M}^{(l,1)}, ... \mathbf{M}^{(l,C)}: \mathbf{M}^{(l,c)} = S(\phi,\mathbf{\widetilde{Q}}^{(l,c)})  \right \}.
\end{equation}


Since the class representative activations are different for each class, a mapping index is obtained for each class-representative activation separately. As shown in Figure~\ref{fig:1}, each mapping index shows the target location for the parameters of the layer $l$ according to the SOFM trained with ($\mathbb{Q}^{(l)}$). Note that in Figure~\ref{fig:1}, a different color represents a different target location. A reorganization process is then performed on each of the parameters of the layer $l$, by simply reorganizing the parameters ($\mathbf{\widetilde{{W}}}^{(l)}$) in the output channel dimension of the layer parameters as follows:


\begin{equation}
    \mathbf{\widetilde{{W}'}}^{(l,c)}_{:,j,:,:} =   \mathbf{\widetilde{{W}}}^{(l)}_{:,\mathbf{M}^{(l,c)}_{j},:,:},
\end{equation}


\noindent  where $ \mathbf{\widetilde{{W}'}}^{(l,c)}$ is the reorganized version of the parameter $ \mathbf{\widetilde{{W}}}^{(l)}$ according to the class-representative activation ($\mathbf{\widetilde{Q}}^{(l,c)}$) of class $c$. It should be noted that, the remapping is also performed to the biases $(\mathbf{B})$ and  batch-norm parameters ($\gamma$ and $\beta$) of the convolutional layer if available. However, since the reorganization process is the same, we only describe the reorganization for the layer weights ($\mathbf{\widetilde{{W}}}^{(l)}$) for simplicity. Since the reorganization is performed on the output layer dimension, the weight parameter of layer $l+1$ should also be reorganized, such that the connection between both layers does not break. The reorganization of the layer $l+1$ can be simply performed by: 


\begin{equation}
    \mathbf{\widetilde{{W}'}}^{(l+1,c)}_{j,:,:,:} =   \mathbf{\widetilde{{W}}}^{(l+1)}_{\mathbf{M}^{(l,c)}_{j},:,:,:}.
\end{equation}

After the reorganization of the parameters at layer $l$ and $l+1$, multiple reorganized versions of the parameters are obtained. In particular, $C$ variants of the reorganized weights are obtained, one corresponding to the mapping index of one class-representative activation. However, since the network should work well on all classes not a particular class, we aggregate the reorganized parameters via a weighted summation. Figure~\ref{fig:1} demonstrates the aggregation process of the reorganized parameters. The weights of the weighted summation are simply  $1/n$, where $n$ is the number of times the values were remapped to the current index. In the illustration in Figure~\ref{fig:1}, $n$ was set to 4 because 4 parameters were mapped to the location shown in red.

%% file: sections/03-03-neuron-rejuvenation.tex
The Final process of the proposed method is rejuvenating the reorganized parameters. As aforementioned, the length of the 1-D map of SOFM $(\phi_i)$ can be less than or equal to the number of output channels $(o)$ of the layer $l$. When the length of $(\phi_i)$ is equal to the number of output channels ($o$), the parameters of $l$ are reorganized into the same number of filters in $l$. In this case, the rejuvenation of the reorganized parameters is done by simply fine-tuning the network. However, when the length of $\phi_i$ is less than $o$, reorganization is coupled with a compression of the parameters into a smaller number of filters. The remaining filters "non-surviving filters" are then reinitialize. We randomly reinitialized the non-surviving parameters, and reset the batch-norm of those filters to zero-mean unit variance. Both the reinitialized non-servicing parameters and the reorganized parameters are rejuvenated by fine-tuning the model with the training data.

%% file: alg/alg_01.tex
\begin{algorithm}[tb]
	\caption{Self-Reorganizing \& Rejuvenating CNN}
	\label{alg:alg1}
	
	\textbf{Input:} $\mathbf{\widetilde{X}}$ \# Training Data \\
	\quad \quad \quad $\mathbf{Y}$ \# Training labels \\
	\quad \quad \quad $f(\mathbf{\widetilde{X}}, \mathbf{\widetilde{W}})$ \# Baseline CNN model \\
	\textbf{Output:} $f(\mathbf{\widetilde{X}}, \mathbf{\widetilde{W'}})$ \# Self-Reorganized \& Rejuvenated CNN
 
	 \While{$n$  \textless $N_{epochs}$}
	 {
	 	\If{n \% r } 
	 	{
    	    Get $\mathbb{Q}^{(l)}$ \# Get class-representative activations of layer $l$\\
    	    $S(\widehat{\phi},\mathbb{Q}^{(l)}) \Leftarrow S(\phi,\mathbb{Q}^{(l)} ) $ \# Train the SOFM  \\
			\For {$c = 1: C$}
			{     	    
			    $ \mathbf{M}^{(c)} = S(\widehat{\phi},\mathbf{Q}^{(l,c)})$ \# Get mapping index for class $c$\\
    	    	$\mathbf{\widetilde{{W}'}}^{(l,c)} \Leftarrow Reorganize(\mathbf{M}^{(c)}, \mathbf{\widetilde{W}}^{(l)} ) $\\
    	    	$\mathbf{\widetilde{{W}'}}^{(l+1,c)} \Leftarrow Reorganize(\mathbf{M}^{(c)}, \mathbf{\widetilde{W}}^{(l+1)} ) $\\
    	    }
		     	   
    	    $\mathbf{\widetilde{{W}}}^{(l)} \Leftarrow aggregate(\mathbf{\widetilde{{W}'}}^{(l,c)})$\\
    	    $\mathbf{\widetilde{{W}}}^{(l+1)} \Leftarrow aggregate(\mathbf{\widetilde{{W}'}}^{(l+1,c)})$\\
        }
        \
	  	Conventional CNN training \\
	  	    
	}
\end{algorithm}

%% file: sections/03-04-training_strategy.tex
The training process for the proposed self-reorganizing and rejuvenating CNN is detailed in Algorithm~\ref{alg:alg1}. For a given network structure  $f(\mathbf{\widetilde{X}}, \mathbf{\widetilde{W}})$, the proposed method can be invoked every $r$ epochs. Every time the proposed method is invoked, the class-representative activations ($\mathbb{Q}^{(l)}$) of layer $l$ are obtained from all the training samples. $\mathbb{Q}^{(l)}$ is then utilized to train a SOFM ($S(\phi,\mathbb{Q}^{(l)})$). The trainable parameters $\mathbf{\widetilde{W}}^{(l)}$ at layer $l$ and  $\mathbf{\widetilde{W}}^{(l+1)}$ layer $l+1$ are then reorganized according to the mapping index attained from the trained SOFM ($S(\widehat{\phi},\mathbb{Q}^{(l)})$). The reorganized weights are then aggregated and assigned back to the original network structure. In the case when the size of the SOFM map is less than the output channels of layer $l$, the remaining parameters are randomly initialized as discussed in the neuron rejuvenation section.

%% file: sections/04-00-exp.tex
%

In this section, we show the results of the proposed method on different network architectures on the CIFAR-100 \cite{cifar100} and Imagenet \cite{imagenet} datasets. When using the CIFAR-100 dataset, all models were trained from scratch to obtain the baseline performance. The training was conducted for 200 epochs. The learning rate was set to 0.1 and reduced by a factor of 5 at epochs 60, 120 and 160. In the first epoch learning rate warm-up was utilized by increasing the learning rate from 0 to 0.1 every iteration to insure model stability \cite{warmup}. When applying the proposed method on CIFAR-100, the self-reorganizing and rejuvenation of convolution layers was performed every 20 epochs ($r=20$). For the Imagenet experiments, pre-trained models from the Pytorch model zoo were utilized as a baseline. The proposed self-reorganizing of convolution layers was then applied once, and the model was left to rejuvenate by end-to-end fine-tuning the self-reorganized model. 

%% file: sections/04-01-exp1_1.tex
To investigate the effect the proposed method has on different convolutional layers the CIFAR-100 dataset was utilized to train multiple baseline architectures. The proposed method was applied to different convolutional layers of each basline model independently. Four baseline models were used, namely; AlexNet \cite{alexnet}, MobileNet \cite{mobilenet}, ResNet-18 and ResNet-50 \cite{resnet}. Some minor changes have been applied to the AlexNet and the ResNet models, in order to adjust them to the CIFAR-100 dataset. In the AlexNet baseline model, local response normalization has been replaced with batch normalization, and the convolution kernels of the first and second layers have been set to $3\times3$ as in \cite{dorefa}. The ResNet models were modified such that the kernel of the first layer has been set to $3\times3$. 

When applying the proposed method, during the neuron reorganization stage, the size of the SOFM was set to 50\% of the output channels of the last convolutional layer. By doing so, all the parameters of the layer are reorganized to half the size of the original layer. The remaining 50\% of the parameters were randomly reinitialized. Table~\ref{Table:1}, shows the results by applying self-reorganization and rejuvenation on four models. For AlexNet, the self-reorganization and rejuvenation was applied independently to convolution layers 3,4 and 5. For the ResNet models, each model was divided into 5 blocks as defined by \cite{resnet}. The class-representative activations were obtained at the output of blocks 3,4 and 5, and self-reorganization and rejuvenation was applied to the last convolution layer of the block.  Similarly, MobileNet layers have been grouped into a stem layer followed by 4 convolution blocks. Similar to the ResNet models, the class-representative features were obtained at the output of the last 3 blocks, and self-reorganization and rejuvenation was applied to the last layer of the block (point-separable convolution layer). The results in Table~\ref{Table:1} show that when the proposed method is applied on higher level layer, the performance gain is more significant. This result was consistent on all models. This can be explained by the fact that lower level layers are required to respond to more simple constructs in abundance such that they are aggregated to higher abstract levels in higher layers. Therefore, clustering in lower layers can be more harmful than it is useful, as can be seen from the results of lower layers shown in Table~\ref{Table:1}. These results are in parallel to the human visual cortex. In the primary visual cortex (V1), neurons fired are distributed across the V1 corresponding to simple constructs in the stimuli \cite{V1,V1IT}. However, activations of neurons start to cluster into specific locations in the inferior temporal (IT) cortex that correspond to certain higher abstraction levels \cite{V1IT,IT}. 

After applying the proposed method onto single independent layers, it is natural to obtain the results from a combination of layers. To that end, the proposed method was applied to layers 4 and 5 of the AlexNet, Resent-18 and Resenet-50, and layers 3 and 4 of MobileNet. These layers were chosen as they have showed performance improvement compared to the corresponding baseline models. For AlexNet, Resent-18 and Resenet-50, after the proposed method was applied to layer 4, the model was frozen until that layer (i.e., layers 1,2,3 and 4 were frozen) and the proposed method was applied to layer 5. For the MobileNet, the proposed method was applied to layer 4 after layer 3 in a similar fashion. The results in Table~\ref{Table:1}. The results show that applying self-reorganization and rejuvenation on multiple layers further improves the performance of the model. Indeed the proposed method increased the utilization of the model capacity.

%% file: tables/table_01.tex
%
%
%

\begin{table}[b]
  \centering
  \caption{Performance in terms of top 1 recognition rate (\%) when the proposed method was applied to different layers of the baseline architectures using the CIFAR-100 dataset . }
  \label{Table:1}
  \resizebox{1.0\columnwidth}{!}{
	\begin{tabular}{c|cccc|c}    
    \toprule
  	\toprule
   	\textbf{Method}						& \textbf{Conv 3} & \textbf{Conv 4}  & \textbf{Conv 5} & \textbf{Conv 4\&5}  & \textbf{Baseline}\\
    \midrule


    AlexNet   		& 65.42 & 66.31 & 67.25 & \textbf{68.38} 	& 64.66 \\    
    ResNet-18  		& 73.79 & 75.04 & 77.10 & \textbf{78.41} 	&74.83 \\    
	ResNet-50  		& 74.81 & 76.33 & 78.75 & \textbf{79.16}	& 76.68 \\

    \midrule
	\textbf{Method}						& \textbf{Conv 2}	& \textbf{Conv 3} & \textbf{Conv 4} & \textbf{Conv 3\&4}  & \textbf{Baseline}\\
    \midrule
    MobileNet 	& 62.95 &	67.61	& 68.65  & \textbf{69.77}	& 66.69\\
    
    \bottomrule
    \bottomrule
    \end{tabular}%
}
\end{table}

%% file: sections/04-01-exp1_2.tex
Different from previous methods \cite{dsd,NR} that utilize pruning and rejuvenation to improve the utilization of the model capacity, the proposed method presented a method for self-reorganizing and rejuvenating the layer parameters according to the similarity of the output activations of that layer. To evaluate the effect of the proposed self-reorganization method on improving the model utilization capacity, a comparative experiment with different pruning based methods was performed. Four over-parameterized baseline models AlexNet \cite{alexnet}, MobileNet \cite{mobilenet}, ResNet-18 and ResNet-50 \cite{resnet} has been utilized with the CIFAR-100 dataset. The models  were modified to fit the CIFAR-100 dataset as described in the previous section.

In the neuron reorganization stage of the proposed method, the size of the SOFM was set to 50\% of the output channels of the last convolutional layer. By doing so, all the parameters of the layer are reorganized to half the size of the original layer. The remaining 50\% of the parameters were randomly reinitialized. The proposed method was only applied to the last convolutional layer of each of the baselines. Two other rejuvenation techniques were also trained. The first model ($\ell_2$), is influanced by the DSD \cite{dsd} approach by relying on $ \|  \mathbf{\widetilde{W}}^{(l)} \|_2$. In particular, we  obtained the $\ell_2$ norm of the weights ($ \|  \mathbf{\widetilde{W}}^{(l)} \|_2$), then the lowest 50\% of the $\ell_2$ norms was reinitialized and rejuvenated by training the model end-to-end. The second method ($\ell_1$) is influenced by \cite{NR}, in the fact that it applies a sparsity constraint on the standard deviation parameter of the batch-norm layer $(\gamma)$.  Then, the lowest 50\% of the $|  \mathbf{\gamma}^{(l)} |$ was reinitialized and rejuvenated by training the model end-to-end. Note that, for a fair comparison the following was applied: (1) in the comparative methods 50\% was reinitialized for a fair comparison with the proposed method. (2) Reorganization and pruning techniques were applied to the same layer in all models to assure consistent comparison between all models. (3) Finally, the re-initialization and rejuvenation of the non-surviving parameters has been the same for all models (as described in the neuron rejuvenation section).

Table ~\ref{Table:2} shows the results of applying the proposed method with comparison to the baseline and compartive models.The results in Table~\ref{Table:2} show that the proposed method outperformed the baseline and the comparative methods.  Note that, since the comparative methods and the proposed method reinitialize the same percentage of the parameters. Yet, the proposed method was able to achieve the most performance gain.  This shows that the performance gain is not due to the re-initialization of the parameters. The proposed method was able to project the layer parameters into a compressed space, then, rejuvenate and extend the capacity of that model improving its performance.

%% file: tables/table_02.tex
%

\begin{table}[t]
  \centering
  \caption{Performance comparison with different baseline architectures on the CIFAR-100 dataset in terms of recognition rate (\%).}
  \label{Table:2}
  \resizebox{1.0\columnwidth}{!}{
	\begin{tabular}{ccc|cc|cc|cc}    
    \toprule
  	\toprule
    							&  \multicolumn{2}{c}{Baseline} 	& \multicolumn{2}{c}{$\ell_2$} & \multicolumn{2}{c}{$\ell_1$} & \multicolumn{2}{c}{Proposed method$_5$}\\ \cmidrule{2-9}
   	Method						& Top 1	  & Top 5	& Top 1	  & Top 5 	& Top 1	  & Top 5 	& Top 1	  & Top 5\\
    \midrule
  
    AlexNet 	& 64.66 & 86.25 & 65.12 & 86.83 & 64.96 & 87.41 & \textbf{67.25} & \textbf{89.97}\\ 
    ResNet-18 	& 74.83 & 92.17 & 74.44 & 92.12 & 74.99 & 92.75 & \textbf{77.10} & \textbf{93.74}\\ 
	ResNet-50 	& 76.68 & 92.69 & 77.27 & 93.45 & 76.54 & 93.38 & \textbf{78.75} & \textbf{94.83}\\ 
	MobileNet 	& 66.69 & 87.38 & 66.28 & 88.02 & 66.49 & 87.96 & \textbf{68.65} & \textbf{88.43}\\
    \bottomrule
    \bottomrule
    \end{tabular}%
}
\end{table}


%% file: tables/table_03.tex

\begin{table}
  \centering
  \caption{Performance comparison with different baseline architectures on the Imagenet dataset in terms of recognition rate (\%).}
  \label{Table:3}
  \resizebox{1.0\columnwidth}{!}{
	\begin{tabular}{ccc|cccc|cc}    
    \toprule
  	\toprule
	&  \multicolumn{2}{c}{\multirow{2}{*}{Baseline}} 	& \multicolumn{2}{c}{\multirow{2}{*}{\parbox{2.5cm}{NR Params \\ \cite{NR}}}}   & \multicolumn{2}{c}{\multirow{2}{*}{Proposed$_5$}}		& \multicolumn{2}{c}{\multirow{2}{*}{Proposed$_{4,5}$}}\\
			  
  \multicolumn{9}{c}{}\\ \cmidrule{2-9}
			  
   	Method			& Top 1	& Top 5		& Top 1	& Top 5 	& Top 1	& Top 5 	& Top 1	& Top 5\\
    \midrule

    ResNet-50 		& 76.15 & 92.87 	& 77.07	& 93.53 	& \textbf{77.56} & \textbf{93.94}	& \textbf{78.22} & \textbf{94.16} \\    
	ResNet-101 		& 77.37 & 93.56 	& 78.78 & 94.24 	& \textbf{79.13} & \textbf{94.63}	& \textbf{80.04} & \textbf{94.98}\\  
	ResNet-152 		& 78.31 & 94.06 	& N/A	& N/A 		& \textbf{80.27} & \textbf{95.11}	& \textbf{81.47} & \textbf{95.34}\\  
	DenseNet-121 	& 74.65 & 92.17 	& 75.50 & 92.51 	& \textbf{76.47} & \textbf{93.86}	& \textbf{77.83} & \textbf{94.53}\\  
	  
    \bottomrule
    \bottomrule
    \end{tabular}%
}
\end{table}


%
%
%
%

%% file: tables/table_04.tex
\begin{table}[b]
  \centering
  \caption{Performance comparison with different baseline architectures on the Imagenet dataset in terms of recognition rate (\%).}
  \label{Table:4}
  \resizebox{0.6\columnwidth}{!}{
	\begin{tabular}{c|c}    
    \toprule
  	\toprule
   	\textbf{Method}				& ResNet-50 Top 1 \\
    \midrule
    Baseline   					& 76.15\\    
    NR Params \cite{NR}  		& 77.07\\    
	RigL$_5$(ERK) \cite{RigL}   & 77.10\\  
	SAC \cite{SAC}				& 76.60\\  

    \midrule
	Proposed$_5$				& \textbf{77.56}\\
	Proposed$_{4,5}$			& \textbf{78.22}\\
    
    \bottomrule
    \bottomrule
    \end{tabular}%
}
\end{table}

%% file: figs/fig2.tex
\begin{figure*}
\begin{center}
	\includegraphics[width=1\textwidth,keepaspectratio]{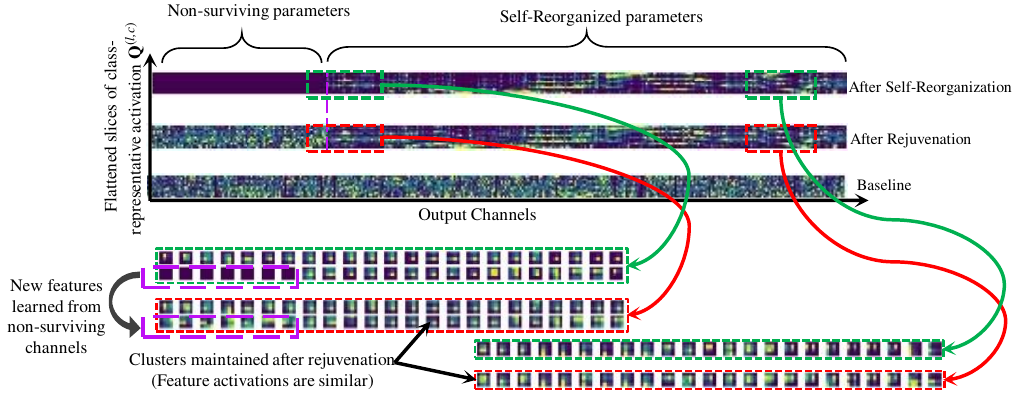}

\end{center}
   \caption{Visualization of the class-representative activation of the MobileNets third layer. class-representative activation of the baseline model, After self-reorganization and after the rejuvenation. Best viewed in color.}
\label{fig:2}
\end{figure*}

%% file: sections/04-02-exp2.tex
In this experiments, we demonstrate the effectiveness of the proposed compared to previous methods. To that end, experiments were done using models on the Imagenet dataset. As baseline models, DenseNet-121 \cite{densenet}, ResNet-50, ResNet-101 and ResNet-152 \cite{resnet} were utilized. For the proposed method, we share the results of two models, the first one is when the proposed method was applied to the fifth convolutional block of that model (i.e, proposed$_5$). The second model is when the proposed model was applied to the last two blocks of the model (i.e. proposed$_{4,5}$). Proposed$_{4,5}$ is achieved by applying the proposed method the 4$^{th}$ layer first. After the proposed method was applied to layer 4, the model was frozen until that layer (i.e., layers 1,2,3 and 4 were frozen) and the proposed method was applied to layer 5. For comparison, the results of neuron rejuvenation when adding a constraint on the number of parameters \cite{NR} were also reported. The results in Table~\ref{Table:3} and Table~\ref{Table:4} show that the proposed method outperform the baseline and previous methods \cite{NR,SAC,RigL}. Moreover, the larger the model the more performance gain was obtained, since larger models tend to be more over-parameterized. A more interesting results could also be seen from Table~\ref{Table:3}. The performance of self-reorganized smaller models becomes comparable with larger models (e.g. self-reorganized and rejuvenated ResNet-101 compared to ResNet-152 baseline). This results can be also seen in Table~\ref{Table:1} on the CIFAR-100 dataset. This testifies that the proposed method better utilizes the capacity of the model. Finally, the results in Table~\ref{Table:1}, Table~\ref{Table:3} and Table~\ref{Table:4} also show that the proposed method is model-agnostic and can work with different baseline models.


%% file: sections/04-03-exp3.tex

In this experiment we visualize the effect of the proposed self-reorganization and rejuvenation on the activations of the layers it was applied to. For the visualization, class-representative activations were utilized. Figure~\ref{fig:2} shows the class-representative features obtained from the the third convolutional layer of MobileNet. The figure shows the class-representative activations from three models, a model that was trained with the conventional method, after the self-reorganization of the parameters and after the parameters were rejuvenated. In the Figure, the each channel of th eclass-representative features was flattened and all 512 channels were displayed. A closeup to some of rejuvenated activations from different channels are also displayed. As can be seen from the figure, the self-reorganization stage projects the parameters into a compressed space, where the activations are compressed according to how similar they are. The neuron rejuvenation stage influences the non-surviving parameters to rejuvenated and learn additional features to supplement the self-reorganized parameters. It is also worth noticing that a rejuvenation stage modifies the reorganized activations as well. These perturbations on the clustered feature manifold explains the performance gain on the same baseline models as discussed in the previous experiments, while simply pruning and rejuvenation did not provide significant gain. Notice also that the rejuvenation does not significantly alter the self-reorganize features and keeps the clustered activations intact.

%% file: sections/05-conclusion.tex
In this paper, we proposed self-reorganizing and rejuvenating CNNs for improving the parameter resource utilization of CNNs. The proposed method draws inspiration from the human visual cortex, where neurons corresponding to certain higher abstraction levels are clustered in specific locations in the inferior temporal (IT) cortex. Similar to the IT cortex, the proposed method clusters deeper layer parameters according to the similarities in the layer activations. By doing so, redundant parameters in those layers are reorganized and aggregated. when rejuvenated, the reorganized parameters have better capability to represent the inputs. Moreover, rejuvenating non-surviving parameters after the self-reorganization learns new feature representations of the input. The rejuvenated parameters can learn additional supplementary features to expand the feature space within the network capacity. Experimental results showed that after the self-reorganization and rejuvenation of higher layers in the network, the baseline network performance is increased in terms of recognition rate. Moreover, the results showed that the proposed method is model-agnostic and can be applied to different CNN architectures without the need for additional operations.

%% file: sections/Appendix-SOFM.tex
Self-organizing feature maps (SOFM) has been introduced by Teuvo Kohonen in \cite{SOM1,SOM2}. SOFM is a type of artificial neural network that is trained using unsupervised learning to produce a low-dimensional, discretized representation (called a map) of the input space of the training samples. SOFM is an on-line stochastic algorithm that updates the weights of the neurons at each step. However, instead of updating weights with error-correction learning, training the SOFM relies on competitive learning. The competitive learning algorithm in SOFMs utilizes a neighborhood function to preserve the topological properties of the input space.

In this work, we utilized the deterministic batch form of the SOFM, which uses all the training data at each step rather than using one sample at a time \cite{batchsom}. Algorithm~\ref{alg:alg2}, shows the batched training process of the SOFM. The goal of the SOFM is to produce a low-dimensional, topology preserving discretized mapping of the input data set $(\mathbb{X} = \{\mathbf{X}(t) | t = 0,1, ... T_f\})$, where $t$ is the index of the sample from the training set and $T_f$ is the index of the end of the current training session (i.e., last sample in the training set). The low-dimensional, topology preserving discretized mapping is a set of neural network nodes $(\mathbb{N} = \{ n_1, ..., n_K \})$ arranges in a grid (typically 2-Dimensional grid). Each node ($n_k$) is associated with a weight vector $\mathbf{\phi}_k(t)$ at time a given time step $t$. To train the SOFM, a best matching unit ($BMU$) is obtained by mapping an input sample $\mathbf{X}(t)$  to a neuron node by:

\begin{equation}
\label{eq:1}
	\begin{split}	
		&BMU(\mathbf{X}(t))= n_k \in \mathbb{N}, \\
		\\
		s.t. \qquad d(\mathbf{X}(t), \mathbf{\phi}_{b}(t)) &\leq d(\mathbf{X}(t), \mathbf{\phi}_k(t)) \qquad \forall \mathbf{\phi}_k(t) \in \mathbf{\Phi},\\
	\end{split}
\end{equation}

where $d(.)$ is the distance function between the training sample $\mathbf{X}(t)$ and a neurons weight $\mathbf{\phi}_k(t)$, and $\mathbf{\phi}_{b}(t)$ is the best matching units weight. Note that, the neurons are arranged in a grid with coordinates representing the neuron location on that grid. Now, the weight vector of the best matching unit is updated to be more similar the training sample. However, to maintain the topology of the SOFM, the neighbors of that neuron are also updated such that they are more similar to the training sample. Remember that, the SOFM is a stochastic algorithm, and all the parameters should be updated after each input. To make sure the weight update only affects the wight vectors of the best matching unit and its neighbors, a neighborhood function $h_{bk}(t)$ is utilized. The parameter update for the SOFM is given by:
 
\begin{equation}
\label{eq:2}
	\mathbf{\phi}_k(t+1)= \mathbf{\phi}_k(t) + \alpha h_{bk}(t)(\mathbf{X}(t) - \mathbf{\phi}_k(t)),
\end{equation}

where $0<\alpha<1$ is a learning rate and $h_{bk}(t)$ is a neighboring function which value decreases for neurons further away from the best matching unit. A commonly used neighborhood function is a Gaussian function:

\begin{equation}
\label{eq:3}
	h_{bk}(t) = exp \left (     \frac{-\left \|  v_b - v_k \right \|}{\sigma(t) }   \right ),
\end{equation}

where $v_b$ and $v_k$ are the coordinates of the best matching unit and node $k$, respectively. $\sigma(t)$ is the variance of the Gaussian function, which decreases each iteration to reduce the area of influenced nodes surrounding the best matching unit. The training is repeated for multiple epochs, until the variance of the neighborhood function $(\sigma(t))$ becomes too small. 

Since all training data is available, the batch formulation of the SOFM \cite{batchsom} replaces the weight update (eq.~\ref{eq:1}), with an update every epoch with the following equation:

\begin{equation}
\label{eq:4}
	\mathbf{\phi}_k(T_f) =     \frac{\sum_{t=0}^{T_f} h_{bk}(t)x(t))}{\sum_{t=0}^{T_f} h_{bk}(t)}.
\end{equation}

In this batch implementation of the SOFM \cite{batchsom}, the parameter updates are done once every epoch, by updating the neighborhood function for each batch and accumulating the numerator and denominator of eq.~\ref{eq:4}. The pseudo code of the utilized SOFM is detailed in Algorithm~\ref{alg:alg2}.

Note that, the result of SOFM is a clustered representation of the training data, that does not require pre-defining the number of clusters. This makes the SOFM a perfect candidate for our purpose, since the number of clusters in the activation of a layer $l$ in the network is unknown. For the proposed method, the input training samples of the SOFM  are the slices of the class-representative activations  $(\widetilde{\mathbf{Q}}^{(l)})$. After training the SOFM with the class-representative activations, we get a clustered representation of the activation of layer $l$. For each class-representative feature $(\widetilde{\mathbf{Q}}^{(l,c)})$ we can obtain a the best matching units for each slice of the class-representative feature $(\widetilde{\mathbf{Q}}_{o,:,:}^{(l,c)})$, which represents the mapping index of that slice on the SOFM feature map. This map is then utilized to reorganize the parameters of layer $l$ as detailed in the original submission manuscript.

%% file: alg/alg_02.tex
\begin{algorithm}
	\caption{Batched Self Organizing Feature Maps}
	\label{alg:alg2}
	Randomly initialize SOFM weights\\
	\While{$epoch$  \textless $epochs_{max}$}
	{
		\For {$t = 0: T_f$}
		{
			get best matching unit (BMU) using eq.~\ref{eq:1}\\
			update learning rate $\alpha$\\
			update neighborhood parameter $\sigma$\\
			get neighborhood function using eq.~\ref{eq:3}\\
			calculate \& accumulate nominator and denominator of eq.~\ref{eq:4}\\
		}
		update SOFM  weights using eq.~\ref{eq:4}\\
	}
\end{algorithm}